\def\year{2020}\relax
\documentclass[letterpaper]{article} 
\usepackage{aaai20}  
\usepackage{times}  
\usepackage{helvet} 
\usepackage{courier}  
\usepackage[hyphens]{url}  
\usepackage{graphicx} 
\usepackage{graphicx}  
\usepackage{latexsym}
\usepackage{color}
\usepackage{amsmath}
\usepackage{hyperref}
\title{Point2Node: Correlation Learning of Dynamic-Node for \\ Point Cloud Feature Modeling}
\author{Wenkai Han,\textsuperscript{\rm 1}
Chenglu Wen,\textsuperscript{\rm 1}\thanks{Corresponding author: Chenglu Wen}
Cheng Wang,\textsuperscript{\rm 1}
Xin Li,\textsuperscript{\rm 2}
Qing Li\textsuperscript{\rm 1}\\
\textsuperscript{\rm 1}School of Informatics, Xiamen University\\ 
422 Siming South Road, Xiamen 361005, China\\
\textsuperscript{\rm 2}School of Electrical Engineering and Computer Science, Louisiana State University\\
Baton Rouge, LA 70803, USA\\
hlxwk0525@gmail.com, clwen@xmu.edu.cn, cwang@xmu.edu.cn, xinli@cct.lsu.edu, hello.qingli@gmail.com 
}
 
\begin{document}

\maketitle

\begin{abstract}
Fully exploring correlation among points in point clouds is essential for their feature modeling. 
This paper presents a novel end-to-end graph model, named Point2Node, to represent a given point cloud. 
Point2Node can dynamically explore correlation among all graph nodes from different levels, and adaptively aggregate the learned features. 
Specifically, 
first, to fully explore the spatial correlation among points for enhanced feature description, in a high-dimensional node graph, we dynamically integrate the node's correlation with self, local, and non-local nodes. 
Second, to more effectively integrate learned features, we design a data-aware gate mechanism to self-adaptively aggregate features at the channel level.
Extensive experiments on various point cloud benchmarks demonstrate that our method outperforms the state-of-the-art. 
\end{abstract}

\section{Introduction}
The recent advance of 3D technologies in autonomous driving, robotics, and reverse engineering has attracted greater attention in understanding and analyzing 3D data. 
Often in the form of point clouds, 3D data is a set of unordered 3D points, sometimes with additional features (e.g., RGB, normal) defined on each point. 
Due to the inherent unorderness and irregularity property of 3D data, 
it is difficult to extend traditional convolutional neural networks to 3D data, 
because a permutation in the order of 3D points should not affect their geometric structures.

Multiple approaches \cite{pointnet,pointnet++,dgcnn,pointweb} have been developed recently to process raw point clouds. 
These approaches can be grouped into several categories according to the point correlation they exploit for feature learning: 
\emph{(1) Self correlation.} 
A seminal work on point cloud feature learning is the PointNet \cite{pointnet}, which uses shared Multi-Layer Perceptions (MLPs) on each point individually, then gathers global representation with pooling functions. Its description ability is limited by not being able to incorporate contextual information from each point's neighbors. 
\emph{(2) Local correlation.} 
PointNet++~\cite{pointnet++} improves the PointNet by applying the PointNet recursively on partitioned pointsets and encoding local features with max pooling. 
PointNet++ essentially treats each point individually and hence still neglects its correlation between neighbors.
Later, DGCNN~\cite{dgcnn} employs shared MLPs to process edges linking the centroid point and its neighboring points. 
But because the edges only connect to the centroid point, the neighboring correlation is not fully exploited. 
The more recent approach, PointWeb \cite{pointweb}, considers a point's correlation by weighting all edges among its neighbors. However, although PointWeb explores the complete correlation from a point's neighbors, it stores redundant connections between two points of different labels.
Furthermore, to our best knowledge, in all these existing approaches, most existing approaches do not consider non-local correlation among distant points that
are not neighbors but could potentially exhibit long-range dependencies. 

Representing point cloud with the graph structure is appropriate for correlation learning. There are some graph-based methods \cite{dynamicedge,dynamicfilters} that use graph convolutions to analyze point cloud on local regions; however, explicit correlations of the neighboring points are not well captured by the predefined convolution operations and this limits their feature modeling capability.

To better leverage correlations of points from different levels, we formulate a novel high-dimensional node graph model, named Point2Node, to learn correlation among 3D points. 
Point2Node proposes a Dynamic Node Correlation module, which constructs a high-dimensional node graph to sequentially explore correlations at different levels.
Specifically, it explores \emph{self correlation} between point's different channels, \emph{local correlation} between neighboring points, and also, \emph{non-local correlation} among distant points having long-range dependency.
Meanwhile, we dynamically update nodes after each correlation learning to enhance node characteristics. 
Furthermore, for aggregating more discriminate features, our Point2Node employs a data-aware gate embedding, named Adaptive Feature Aggregation, to pass or block channels of high-dimensional nodes from different correlation learning, thereby generating new nodes to balance self, local and non-local correlation. 
Experimental results on various datasets show better feature representation capability. Our contributions are summarized as follows:
\begin{itemize}
\item We propose a Point2Node model to dynamically reasons the different-levels correlation including self correlation, local correlation, and non-local correlation. This model largely enhances nodes characteristic from different scale correlation learning, especially embedding long-range correlation.
\item We introduce a data-aware gate mechanism, which self-adaptively aggregates features from different correlation learning and maintains effective components in different levels, resulting in more discriminate features.
\item Our end-to-end Point2Node model achieves state-of-the-art performance on various point cloud benchmark, thus demonstrating its effectiveness and generalization ability.
\end{itemize}

\section{Related work}
\textbf{View-based and volumetric methods.} 
A classic category of deep-learning based 3D representations is to use the multi-view approach \cite{voandmulti,multiview,projective}, where a point cloud is projected to a collection of images rendered from different views. By this means, point data can be recognized using conventional convolution in 2D images. 
However, these viewed-based methods only achieve a dominant performance in classification. Scene segmentation is nontrivial due to the loss of the discriminate geometric relationship of the 3D points during 2D projection. 

Influenced by the success of traditional convolution in regular data format (e.g., 2D image), another strategy is to generalize CNN to 3D voxel structures~\cite{voxnet,modelnet,scannet}, then assign 3D points to regular volumetric occupancy grids. Due to the intractable computational cost, these methods are usually constrained by low spatial resolution. To overcome this limitation, Kd-Net \cite{kdnet} and OctNet \cite{ocnet} skip the computation on empty voxels so that they can handle higher-resolutional grids. However, during voxelization, it is inevitable to lose geometric information. To avoid geometric information loss, unlike view-based and volumetric methods, our method processes 3D point clouds directly.

\textbf{Point cloud-based methods.} The pioneering method in this category is PointNet \cite{pointnet}, which considers each point independently with shared MLPs for permutation invariance and aggregates the global signature with max-pooling. 
PointNet is limited by the fine-grained information in local regions. PointNet++ \cite{pointnet++} extends PointNet by exploiting a hierarchical structure to incorporate local dependencies with Farthest Point Sampling (FPS) and a geometric grouping strategy. 
DGCNN \cite{dgcnn} proposes an edge-conv operation which encodes the edges between the center point and k nearest neighbors with shared MLPs. PointWeb \cite{pointweb} enriches point features by weighting all edges in local regions with point-pair difference. Our method also exploits local edges, but we explore the adaptive connection of similar nodes in local regions, to avoid redundant connections between two points of different labels, and this leads to  better shape-awareness.

A few other methods extend the convolution operator to process unordered points. PCCN \cite{PCCN} proposes parametric continuous convolutions that exploit the parameterized kernel functions on point clouds. By reordering and weighting local points with an $\chi$-transformation, PointCNN \cite{PointCNN} exploits the canonical order of points and then applies convolution. PointConv \cite{pointconv} extends the dynamic filter to a new convolution operation and build deep convolutional networks for point clouds. A-CNN \cite{A-CNN} employs annular convolution to capture the local neighborhood geometry of each point. Unlike these methods that use convolution in point clouds, we pay our attention to correlation learning among 3D points in point clouds.

\textbf{Graph-based methods.} Because it is flexible to represent irregular point clouds with graphs, there is an increasing interest in generalizing convolutions to the graphs. In line with the domain in which convolutions operate, graph-based methods can be categorized into spectral and non-spectral methods. Spectral methods \cite{semisuper,syncspecnn} perform convolution in the spectral representation, which is constructed by computing with a Laplacian matrix. The major limitation of these methods is that they require a fixed graph structure. Thus they have difficulty in adapting to point clouds that have varying graph structures. Instead, we perform correlation learning on dynamic nodes, which are not limited by varying graph structures. Alternatively, non-spectral methods \cite{dynamicedge,diffusion,dynamicfilters} perform convolution-like operations on the spatially close nodes of a graph, thereby propagating information along the neighboring nodes. But, hand-crafted ways of encoding local contextual information limits their feature modeling capability; instead, for more fine-grained contextual information in the local region, we explore the direct correlation between local nodes.

\begin{figure*}[htbp]
    \centering
    \includegraphics[width=2.1\columnwidth]{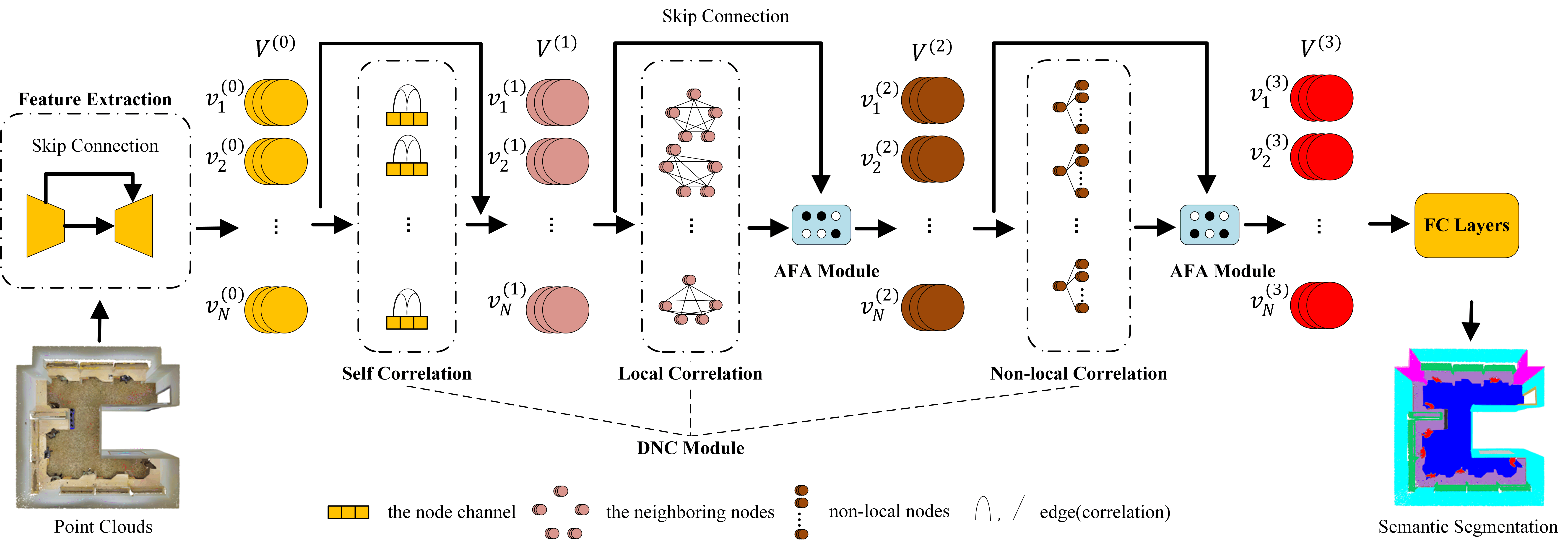}
    \caption{Our Point2Node architecture. First, Point2Node converts input points into high-level representation through Feature Extraction, then constructs high-dimension nodes $V^{(0)}$. Second, the Dynamic Node Correlation (DNC) module (which consists of self correlation, local correlation and non-local correlation) and the Adaptive Feature Aggregation (AFA) module are employed to enhance the node characteristics, resulting in different nodes (different colors). Finally, the nodes are assigned different labels through FC Layers.
    }
    \label{framework}
\end{figure*}

\section{Method}

\subsection{Overview}
We illustrate our end-to-end Point2Node framework in Fig. \ref{framework}.
Given a set of unordered points $P=\{p_i \in R^{3+d},i=1,2,...,N\}$, where $N$ is the number of points and $3+d$ denotes the $xyz$-dimension and additional properties (e.g., $d=3$ for RGB or normal information), through \emph{Feature Extraction}, we construct their high-level representation, forming high-dimensional nodes $V^{(0)}=\{v^{(0)}_i \in R^C,i=1,2,...,N\}$, where each node $v^{(0)}_i$ is a learned multi-channel feature vector of  point $p_i$, and $C$ is the dimension of each node.
The \emph{Feature Extraction} module employs a $\chi$-conv operator from PointCNN \cite{PointCNN} and builds a U-Net~\cite{U-net} structure for feature learning. 
Then through a \emph{Dynamic Node Correlation} (DNC) module, we explore point correlations from different levels, and dynamically generate high-dimensional nodes $V^{(1)}$, $V^{(2)}$ and $V^{(3)}$, respectively. 
Simultaneously, we build an \emph{Adaptive Feature Aggregation} (AFA) module to self-adaptively aggregate the features of nodes learned in different levels. 
Finally, the high-dimensional nodes are fed into fully-connected layers for label assignment.

\subsection{Dynamic Nodes Correlation Module}
To fully explore correlation among the points from different levels, we propose a \emph{Dynamic Nodes Correlation} module. 
Given the initial high-dimensional nodes $V^{(0)}=\{v^{(0)}_i \in R^C,i=1,2,...,N\}$, we introduce the \emph{cross-channel} edges and \emph{cross-node} edges. 
\emph{Cross-channel edges} are defined between two channels of the same node, and they are used to explore correlation between this node's different channels.  
\emph{Cross-node edges} are defined between different nodes, and have two types.
Between neighboring (spatially close) nodes, \emph{local edges} are defined to explore local contextual information;
while between non-local (spatially distant) nodes, \emph{non-local edges} are defined to exploit latent long-range dependencies. 
Therefore, three types of correlation: \emph{Self Correlation}, \emph{Local Correlation}, and \emph{Non-local Correlation} will be learned in this DNC module. 
Meanwhile, we dynamically update the high-dimensional nodes after each correlation learning. 
The learning of these three types of correlation is elaborated in the following sections.

\subsection{Self Correlation}
The dependencies among channels of the same node are denoted by \emph{self correlation}. 
Such dependencies are informative for feature representation. 
To model self correlation of each node, we can construct edges between different channels. 
However, for each channel, we do not compute its correlation weights w.r.t. other channels, but make a simplification. 
We use convolution to integrate these weights from different channels and encode them into a single weight. 

\textbf{Channel Weights.} Inspired by traditional 1D convolution that aggregates all channels of one pixel in a 2D image, we similarly do such an encoding to integrate each channel's correlations into one weight through Multi-Layer Perception (MLP).
To reduce the number of parameters, we implement the MLP with 1D separable convolutions \cite{xception}. 
Consequently, for each $C$-dimensional node $v^{(0)}_i \in V^{(0)}$, we can calculate a $C$-dimensional channel weight $w_i \in R^C$ 
\begin{equation}
    w_i = MLP(v^{(0)}_i).
\label{eq1}
\end{equation}
And to make weight coefficients comparable between different channels, we perform a normalization using $softmax$, 
\begin{equation}
    \overline{w_{i,c}}=\frac{exp(w_{i,c})}{\sum_{j=1}^{C}exp(w_{i,j})},
    \label{eq2}
\end{equation}
where $w_{i,c}$ is the $c$-th component of $w_i$, and $\overline{w_{i,c}}$ is the normalized $w_{i,c}$.

\textbf{Updating Nodes.} After correlation learning, we update nodes with residual connections: 
\begin{equation}
    v^{(1)}_i = v^{(0)}_i \oplus \alpha (\overline{w_{i}} \odot v^{(0)}_i),
    \label{eq3}
\end{equation}
where $\oplus$ and $\odot$ denote channel-wise summation and multiplication, respectively, and $\alpha$ denotes a learnable scale parameter. 
This update transforms nodes $V^{(0)}$ into new nodes $V^{(1)}=\{v^{(1)}_i \in R^C,i=1,2,...,N\}$.

Note that unlike directly transforming $V^{(0)}$ to $V^{(1)}$ by convolutions, which share weights among all the nodes, our \emph{self correlation} module learns different correlations in different nodes and can embody different nodes' individual characteristics.

\subsection{Local Correlation}
To fully capture local contextual information, which is of central importance in point cloud understanding, we also build a new \emph{Local Correlation} sub-module.
Existing approaches~\cite{dgcnn,pointweb} usually construct partial or complete edges of local regions for contextual information, but these tend to store redundant connections between two points of different labels.
In contrast, our sub-module adaptively connects locally similar nodes to construct a local graph, 
whose adjacency matrix weights correlation of relevant node pairs. 

\textbf{Local Graph.} Given $V^{(1)}=\{v^{(1)}_i \in R^C,i=1,2,...,N\}$, we consider each node as a centroid and construct a local graph. 
To increase the receptive field without redundant computational cost, we employ the dilated KNN \cite{PointCNN} to search for neighboring nodes, and form \emph{Dilated $K$-Nearest Neighbor Graphs} (DKNNG) $G^{(l)}=\{G^{(l)}_i \in R^{K \times C},i=1,2,...,N\}$ in local regions. Specifically, we sample $K$ nodes at equal intervals from the top $K \times d$ neighbor nodes, where $K$ denotes the number of samples, and $d$ is the dilate rate.

\textbf{Adjacency Matrix.} For local correlation learning of node $v^{(1)}_i$, given its DKNNG $G^{(l)}_i$ where $V^{(l)}_i=\{V^{(l)}_i(k) \in R^C,k=1,2,...,K\}$ is the neighboring nodes set of centroid node $v^{(1)}_i$, we construct its adjacency matrix $m^{(l)}_i \in R^{K \times K}$ for all the directed edges in $G^{(l)}_i$. 
Inspired by the Non-Local Neural Network ~\cite{non-local}, which has been proven to be effective in capturing pixel correlation in 2D images, 
we generalize its non-local operation to our graph structure, to generate the adjacency matrix.
Please refer to Appendix A for more detail about our generalization.

Specifically, we first map $V^{(l)}_i$ to low-dimension feature space with two 1-D convolutional mapping functions, $\theta^{(l)}$ and $\varphi^{(l)}$, resulting in low-dimension node representations $Q^{(l)}_i \in R^{K\times \frac{C}{r}}$ and $X^{(l)}_i \in R^{K\times \frac{C}{r}}$ respectively, 
\begin{equation}
    Q^{(l)}_i=\theta^{(l)}(V^{(l)}_i),X^{(l)}_i=\varphi^{(l)}(V^{(l)}_i),
    \label{eq4}
\end{equation}
where $r$ is the reduction rate and is a shared super-parameter in the full paper. 
With $Q^{(l)}_i$ and $X^{(l)}_i$, we compute $m^{(l)}_i$ by matrix multiplication, 
\begin{equation}
    m^{(l)}_i=Q^{(l)}_i \otimes (X^{(l)}_i)^T,
    \label{eq5}
\end{equation}
where $\otimes$ denotes matrix multiplication.

Next, we prune the adjacency matrix $m^{(l)}_i$ with the $softmax$ function (Eq.\ref{eq6}) to cut off redundant connections that link nodes of different labels, resulting in $\overline{m^{(l)}_i}$.
\begin{equation}
    \overline{m^{(l)}_i}(x,y)=\frac{exp({m^{(l)}_i}(x,y))}{\sum_{k=1}^{K}exp({m^{(l)}_i}(x,k))},
    \label{eq6}
\end{equation}
where $\overline{m^{(l)}_i}(x,y)$ is the masked ${m^{(l)}_i}(x,y)$, and the final weight of the edge that links the $x$-th node and $y$-th node in the neighboring nodes set $V^{(l)}_i$.

\textbf{Update Nodes.} We simultaneously update the neighboring nodes $V^{(l)}_i$ using the correlation function $\overline{m^{(l)}_i}$ (Eq.~\ref{eq7})
Finally, we encode learned local contextual information to the centroid node with $Max\_Pooling$ function (Eq.~\ref{eq8}). 
\begin{equation}
    {V^{(l)}_i}_{up}=\overline{m^{(l)}_i} \otimes V^{(l)}_i.
    \label{eq7}
\end{equation}
\begin{equation}
    {v^{(1)}_{i}}_{up}=Max\_Pooling({V^{(l)}_{i}}_{up}),
    \label{eq8}
\end{equation}
where ${V^{(l)}_i}_{up}$ and ${v^{(1)}_{i}}_{up}$ are updated neighboring nodes and centroid node, respectively.

\subsection{Non-local Correlation}

To better exploit latent long-range dependency, which is essential in capturing global characteristics of point clouds, we also propose a \emph{Non-local Correlation} sub-module. 
Traditional point cloud networks usually construct global representations by stacking multi-layers local operations, 
but such a high-level global representation can hardly capture direct correlation between spatially distant nodes. 
To model such non-local correlation, our proposed module adaptively connects non-local nodes to construct a global graph, whose adjacency matrix reveals latent long-range dependencies between nodes. 

\textbf{Adjacency Matrix.} 
Given nodes $V^{(2)}=\{v^{(2)}_i \in R^C,i=1,2,...,N\}$, we construct a global graph $G^{(g)}$ and an adjacency matrix $m^{(g)} \in R^{N \times N}$. $m^{(g)}$ is constructed in a similar way to \emph{local correlation}. 
We construct the low-dimensional $Q^{(g)} \in R^{N\times \frac{C}{r}}$ and $X^{(g)} \in R^{N\times \frac{C}{r}}$ from mapping functions $\theta^{(g)}$ and $\varphi^{(g)}$, 
\begin{equation}
    Q^{(g)}=\theta^{(g)}(V^{(2)}), X^{(g)}=\varphi^{(g)}(V^{(2)}),
    \label{eq9}
\end{equation}
then compose the adjacency matrix by
\begin{equation}
    m^{(g)}=Q^{(g)} \otimes (X^{(g)})^T.
    \label{eq10}
\end{equation}

Unlike \emph{local correlation}, the non-local correlation learning can explore dependence between nodes that have different features in different scales. Therefore, to make weight coefficients comparable, we perform a normalization on the adjacency matrix $\overline{m^{(g)}}$,
\begin{equation}
    \overline{m^{(g)}}(x,y)=\frac{exp({m^{(g)}}(x,y))}{\sum_{k=1}^{N}exp({m^{(g)}}(x,k))},
    \label{eq11}
\end{equation}
where $\overline{m^{(g)}}(x,y)$ is the normalized weight of the edge linking nodes $x$ and $y$.

\textbf{Updating Nodes.} We update all the nodes $V^{(2)}$ with the correlation function $\overline{m^{(g)}}$, resulting in ${V^{(2)}}_{up}$: 
\begin{equation}
    {V^{(2)}}_{up}=\overline{m^{(g)}} \otimes V^{(2)}.
    \label{eq12}
\end{equation}

Note that all the three aforementioned sub-modules perform node-wise operations. These operations are permutation-equivalent (See a proof in Appendix B) and hence can effectively process unordered 3D points.

\subsection{Adaptive Feature Aggregation}
Traditional neural networks aggregate different features using linear aggregation such as skip connections. 
But linear aggregations simply mix up channels of the features and are not data-adaptive. 
To better reflect the characteristics of given point cloud and make the aggregation data-aware, 
we propose an \emph{Adaptive Feature Aggregation} (AFA) module~\footnote{Note that because the \emph{self correlation} sub-module already enhances the cross-channel attention by linear aggregation, we do not place an AFA after \emph{self correlation}, as Fig. \ref{framework} shows.
} and use a gate mechanism to integrate node features according to their characteristics.
To model the channel-wise characteristics and their relative importance, we consider two models, namely, the parameter-free to parameterized descriptions. 

\textbf{Parameter-free Characteristic Modeling.} Given high-dimensional nodes $V^{(1)}$ and ${V^{(1)}}_{up}$ that are from different correlation learning, we first describe their characteristics using a global node average pooling, and obtain channel-wise descriptors $s_1 \in R^{1 \times C}$ and $s_2 \in R^{1 \times C}$:
\begin{equation}
    s_{1,c}=\frac{1}{N}\sum_{i=1}^{N}v^{(1)}_{i,c}, \; s_{2,c}=\frac{1}{N}\sum_{i=1}^{N}{v^{(1)}_{i,c}}_{up},
    \label{eq13}
\end{equation}
where $s_{t,c}$ refers to the $c$-th channel of $s_t$, $v^{(1)}_{i,c}$ denotes the $c$-th channel of the $i$-th node $v^{(1)}_{i}$ in $V^{(1)}$, and ${v^{(1)}_{i,c}}_{up}$ denotes the $c$-th channel of the $i$-th node ${v^{(1)}_{i}}_{up}$ in ${V^{(1)}}_{up}$.

\textbf{Parameterized Characteristic Modeling.} 
To further explore the characteristics of the high-dimensional nodes, we introduce learnable parameters into characteristics. Specifically, we employ different MLPs to transform channel-wise descriptors $s_1$ and $s_2$ to new descriptors $z_1 \in R^{1 \times C}$ and $z_2 \in R^{1 \times C}$. To reduce computational costs, we simplify the MLPs through one dimension-reduction convolution and one dimension-increasing convolution layers:
\begin{equation}
    z_1=MLP_1(s_1), \; z_2=MLP_2(s_2).
    \label{eq14}
\end{equation}

\begin{figure}[htbp]
    \centering
    \includegraphics[width=.95\columnwidth]{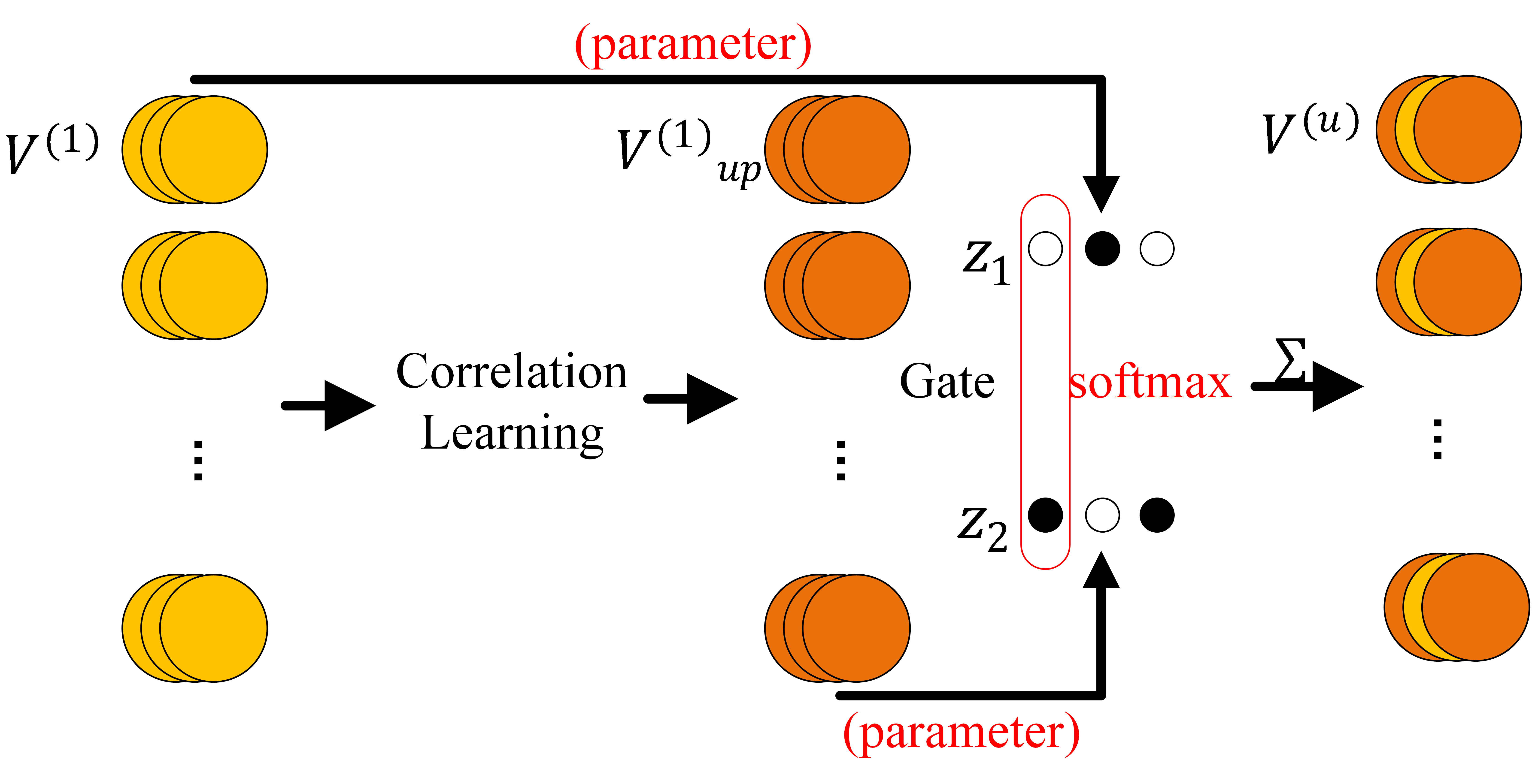}
    \caption{Illustration of Adaptive Feature Aggregation.}
    \label{AA}
\end{figure}

\textbf{Gate Mechanism.} 
Informative descriptions of the aggregated nodes features consist of discriminate correlation representations of different levels. 
Linear aggregations mix up different correlation representations. 
To combine features in a non-linear and data-adaptive way, we construct a gate mechanism (Fig.~\ref{AA}) by applying a $softmax$ on the compact features $z_1$ and $z_2$, by
\begin{footnotesize}
\begin{equation}
    m_{1,c}=\frac{exp(z_{1,c})}{exp(z_{1,c})+exp(z_{2,c})},m_{2,c}=\frac{exp(z_{2,c})}{exp(z_{1,c})+exp(z_{2,c})},
    \label{eq15}
\end{equation}
\end{footnotesize}
where $m_{1,c}$ and $m_{2,c}$ denote the masked $z_1$ and $z_2$ on the $c$-th channel, and $m_{1,c}+m_{2,c}=1$.

\textbf{Aggregation.} Finally, we aggregate the ultimate nodes $V^{(u)}=\{V^{(u)}_{1:N,c} \in R^N,c=1,2,...,C\}$ by the masked $m_1$ and $m_2$, 
\begin{equation}
    V^{(u)}_{1:N,c}=m_{1,c}\cdot V^{(1)}_{1:N,c}+m_{2,c}\cdot {V^{(1)}_{1:N,c}}_{up},
    \label{eq16}
\end{equation}
where $V^{(u)}_{1:N,c}$ denotes the $c$-th channel set of $V^{(u)}$.

This AFA module performs node-wise operations, which are again permutation-equivariant (A proof is given in Appendix C) and suitable for unordered point clouds.

\section{Experiments}
We evaluate our model on multiple point cloud benchmarks, including Stanford Large-Scale 3D Indoor Space (S3DIS) \cite{s3dis} for semantic segmentation, ScanNet \cite{scannet} for semantic voxel labeling, and ModelNet40 \cite{modelnet} for shape classification.

\subsection{S3DIS Semantic Segmentation}
\begin{table*}[ht]
\centering
\resizebox{2.1\columnwidth}{!}{
\begin{tabular}{c|c c c|c c c c c c c c c c c c c c}
   \hline
   Methods & OA & mAcc & mIoU & {ceiling} & {flooring} & {wall} & {beam} & {column} & {window} & {door} & {table} & {chair} & {sofa} & {bookcase} & {board} & {clutter}\\
   \hline
   PointNet   & 78.50 & 66.20 & 47.80 & 88.00 & 88.70 & 69.30 & 42.40 & 23.10 & 47.50 &  51.60 & 54.10 & 42.00 & 9.60 & 38.20 & 29.40 & 35.20 \\
   SPGraph    & 85.50 & 73.00 & 62.10 & 89.90 & 95.10 & 76.40 & {\bfseries 62.80} & 47.10 & 55.30 & {\bfseries 68.40} & 69.20 & {\bfseries 73.50} & 45.90 & {\bfseries 63.20} & 8.70 & 52.90 \\
   RSNet      & -     & 66.45 & 56.47 & 92.48 & 92.83 & 78.56 & 32.75 & 34.37 & 51.62 & 68.11 & 59.72 & 60.13 & 16.42 & 50.22 & 44.85 & 52.03 \\
   DGCNN      & 84.10 & -     & 56.10 & -     & -     & -     & -     & -     & -     & -     & -     & -     & -     & -     & -     & -     \\
   PointCNN   & 88.14 & 75.61 & 65.39 & {\bfseries 94.80} & {\bfseries 97.30} & 75.80 & 63.30 & 51.70 & 58.40 & 57.20 & 69.10 & 71.60 & 61.20 & 39.10 & 52.20 & 58.60 \\
   PointWeb   & 87.31 & 76.19 & 66.73 & 93.54 & 94.21 & 80.84 & 52.44 & 41.33 & 64.89 & 68.13 & 71.35 & 67.05 & 50.34 & 62.68 & 62.20 & 58.49 \\
   \hline
   Point2Node & {\bfseries 89.01} & {\bfseries 79.10} & {\bfseries 70.00} & 94.08 & 97.28 & {\bfseries 83.42} & 62.68 & {\bfseries 52.28} & {\bfseries 72.31} & 64.30 & {\bfseries 75.77} & 70.78 & {\bfseries 65.73} & 49.83 & 60.26 & {\bfseries 60.90} \\
   \hline
   
\end{tabular}
}
\caption{Semantic segmentation results on S3DIS dataset with 6-folds cross validation.}
\label{s3dis1}
\end{table*}

\begin{table*}[ht]
\centering
\resizebox{2.1\columnwidth}{!}{
\begin{tabular}{c|c c c|c c c c c c c c c c c c c c}
   \hline
   Methods & OA & mAcc & mIoU & {ceiling} & {flooring} & {wall} & {beam} & {column} & {window} & {door} & {table} & {chair} & {sofa} & {bookcase} & {board} & {clutter}\\
   \hline
   PointNet   & -     & 48.98 & 41.09 & 88.80 & 97.33 & 69.80 & 0.05 & 3.92 & 46.26 & 10.76 & 58.93 & 52.61 & 5.85 & 40.28 & 26.38 & 33.22 \\
   SegCloud   & -     & 57.35 & 48.92 & 90.06 & 96.05 & 69.86 & 0.00 & 18.37 & 38.35 & 23.12 & 70.40 & 75.89 & 40.88 & 58.42 & 12.96 & 41.60 \\
   SPGraph    & 86.38 & 66.50 & 58.04 & 89.35 & 96.87 & 78.12 & 0.00 & {\bfseries 42.81} & 48.93 & 61.58 & 84.66 & 75.41 & 69.84 & 52.60 & 2.10 & 52.22 \\
   PointCNN   & 85.91 & 63.86 & 57.26 & 92.31 & 98.24 & 79.41 & 0.00 & 17.60 & 22.77 & 62.09 & 74.39 & 80.59 & 31.67 & 66.67 & 62.05 & 56.74 \\
   PCCN       & -     & 67.01 & 58.27 & 92.26 & 96.20 & 75.89 & {\bfseries 0.27} & 5.98 & {\bfseries 69.49} & {\bfseries 63.45} & 66.87 & 65.63 & 47.28 & 68.91 & 59.10 & 46.22 \\
   PointWeb   & 86.97 & 66.64 & 60.28 & 91.95 & {\bfseries 98.48} & 79.39 & 0.00 & 21.11 & 59.72 & 34.81 & 76.33 & {\bfseries 88.27} & 46.89 & 69.30 & 64.91 & 52.46 \\
   GACNet     & 87.79 & -  & 62.85 & 92.28 & 98.27 & 81.90 & 0.00 & 20.35 & 59.07 & 40.85 & {\bfseries 85.80} & 78.54 & {\bfseries 70.75} & 61.70 & {\bfseries 74.66} & 52.82 \\
   \hline
   Point2Node & {\bfseries 88.81} & {\bfseries 70.02} & {\bfseries 62.96} & {\bfseries 93.88} & 98.26 & {\bfseries 83.30} & 0.00 & 35.65 & 55.31 & 58.78 & 79.51 & 84.67 & 44.07 & {\bfseries 71.13} & 58.72 & {\bfseries 55.17} \\
   \hline
\end{tabular}
}
\caption{Semantic segmentation results on S3DIS dataset evaluated on Area 5.}
\label{s3dis2}
\end{table*}

\textbf{Dataset and Metric.} The S3DIS \cite{s3dis} dataset contains six sets of point cloud data from three different buildings (including 271 rooms). Each point, with xyz coordinates and RGB features, is annotated with one semantic label from 13 categories. We followed the same data pre-processing approaches given in PointCNN \cite{PointCNN}.

We conducted our experiments in 6-fold cross validation \cite{pointnet} and fifth fold cross validation \cite{segcloud}. Because there are no overlaps between Area 5 and the other areas, experiments on Area 5 could measure the framework's generalizability. For the evaluation metrics, we calculated mean of class-wise intersection over union (mIoU), mean of class-wise accuracy (mAcc), and overall point-wise accuracy (OA).

\begin{figure*}[htbp]
    \centering
    \includegraphics[width=2.1\columnwidth]{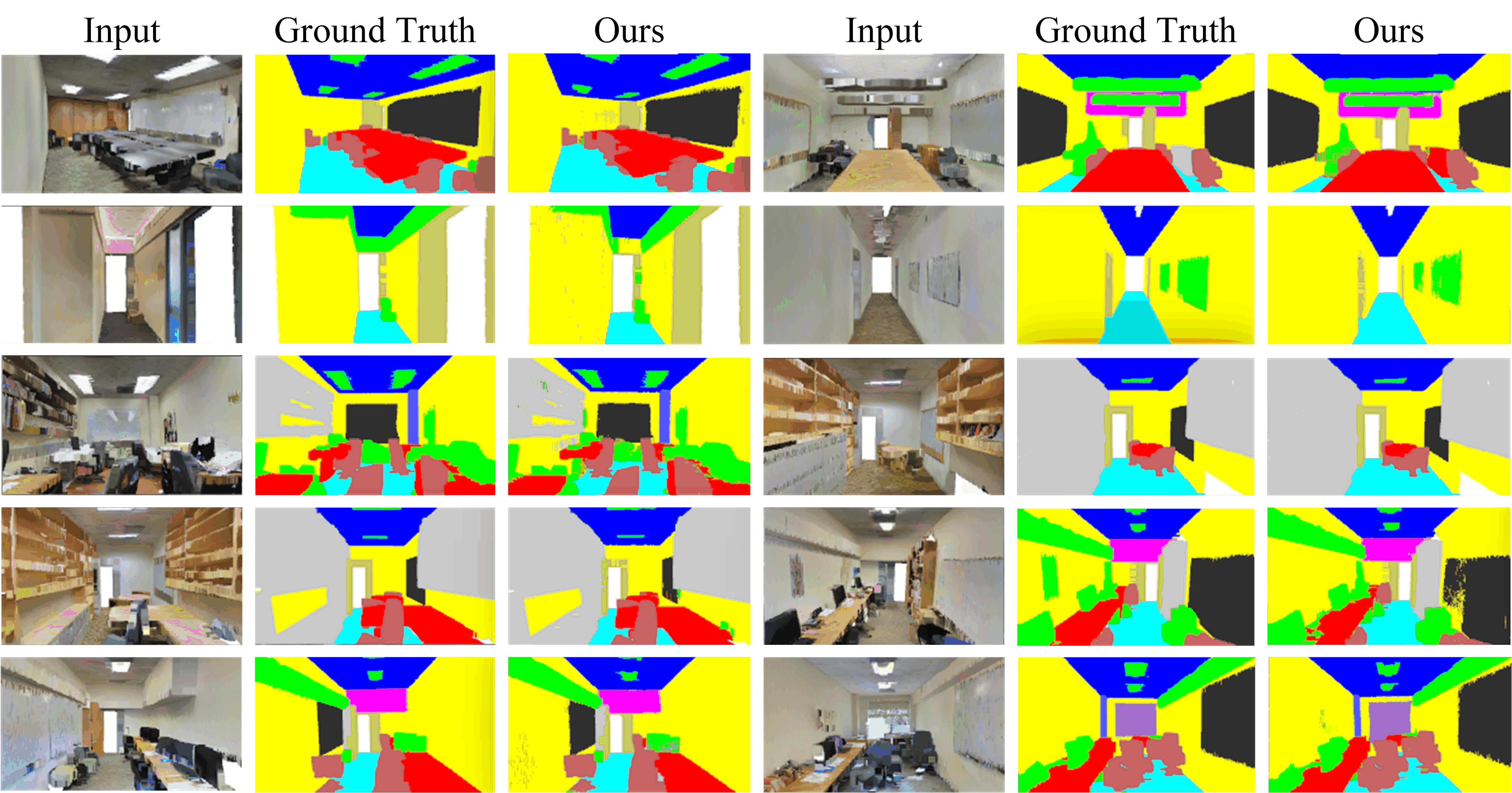}
    \caption{Visualization of semantic segmentation results on S3DIS dataset.}
    \label{visual s3dis}
\end{figure*}

\textbf{Performance Comparison.} The quantitative results are given in Table \ref{s3dis1} and Table \ref{s3dis2}. Our Point2Node performs better than the other competitive methods on both evaluation settings. In particular, despite the dataset contains objects of various sizes, our approach achieves considerable gains on both relatively small objects (e.g., chair, table, and board) and big structures (e.g., ceiling, wall, column and bookcase), which demonstrates that our model can capture discriminate features on both local and global level. For some similar geometric structures such as column vs. wall, our model also works very well, which indicates that our model is capable of distinguishing ambiguous structures due to the latent correlation they capture between the structures and other structures of the scene. Some qualitative results on different scenes are visualized in Fig. \ref{visual s3dis}.

\subsection{Ablation Study}
For ablation studies, we integrate each correlation learning sub-module and its corresponding feature aggregation methods, forming \emph{self correlation} block, \emph{local correlation} block and \emph{non-local correlation} block, respectively. Our baseline method employs $\chi$-conv \cite{PointCNN} and builds Encoder-Decoder structure without any correlation learning block. We conducted ablation studies on Area 5 of the S3DIS.

To show the effectiveness of each proposed block in our model, we separately added each block after the Encoder-Decoder structure while keeping the rest unchanged in our baseline, forming Self only, Local only and Non-local only. Our proposed Point2Node employs all three blocks to form a more powerful framework. As shown in Table \ref{abstudy1}, each block plays an important role in our framework. Moreover, compared with the baseline, the performance improvement (6.26\%) of Point2Node exceeds the simple summation (5.02\%) of the separate block, demonstrating that our sequential structure further enhances the node features.

\begin{table}[ht]
\centering
\begin{tabular}{c|c c}
   \hline
   Ablation studies & mIoU  & $\Delta$ \\
   \hline
   Baseline          & 56.70             & +0.00\\
   Self only         & 57.45             & +0.75\\
   Local only        & 57.91             & +1.23\\
   Non-local only    & 59.74             & +3.04\\
   Parallel\_1       & 59.51             & +2.81\\
   Parallel\_2       & 60.57             & +3.87\\
   LA                & 58.17             & +1.47\\
   Parameter-free AA & 60.57             & +3.87\\
   Point2Node        & {\bfseries 62.96} & {\bfseries +6.26} \\
   \hline
\end{tabular}
\caption{Ablation studies on the S3DIS dataset evaluated on Area 5.}
\label{abstudy1}
\end{table}

To study the effectiveness of sequentially dynamic nodes update strategy in our model, we compared our original sequential structure with parallel structure variants. Specifically, (1) by placing three blocks in parallel, denoted by ``Parallel\_1'', the nodes get updated just once after these parallel blocks; (2) by placing the \emph{local correlation} block and \emph{non-local correlation} block in parallel, denoted by ``Parallel\_2'', the nodes get updated after \emph{self correlation} and after the parallel blocks. 
As shown in Table \ref{abstudy1}, ``Parallel\_2'' performs better than ``Parallel\_1''. The proposed Point2Node model, with sequential and fully updated nodes, has the best performance and indicates the effectiveness of 
this node update mechanism. 

To show our {adaptive aggregation} can lead to more discriminate features than {linear aggregation}, we replaced the adaptive aggregation of \emph{local correlation} block and \emph{non-local correlation} block with the {linear aggregation} in the architecture of Point2Node, to form a new network (LA). 
For a fair comparison, we also employed the parameter-free characteristics of the AFA module to construct a parameter-free adaptive aggregation, named ``Parameter-free AA''.  
The results demonstrate that our adaptive aggregation captures more discriminate features than linear aggregation does, and the parameterized adaptive aggregation (Point2Node) is more effective than parameter-free AA.
\begin{table}[ht]
\centering
\begin{tabular}{c|c c}
    \hline
                     & model size       & mIoU\\
    \hline
    Baseline         & {\bfseries 10.98M} & 56.70\\
    Point2Node       & 11.14M(+1.46\%)    &{\bfseries 62.96(+11.04\%)}\\
    \hline
\end{tabular}
\caption{Comparisons results on model complexity.}
\label{model size}
\end{table}

As shown in Table \ref{model size}, we significantly improve the performance (+11.03\%) while increasing very few parameters (+1.46\%).  In our Point2Node, we frequently employ different MLPs, and we control the reduction rate $r$ ($r=8$ in our experiments) to reduce parameters in adjacency matrix computing and parameterized characteristics.

\begin{table}[htbp]
\centering
\begin{tabular}{l|c}
\hline
Mehtods    & OA \\
\hline
3DCNN \cite{3dcnn}      & 73.0 \\
PointNet \cite{pointnet}   & 73.9\\
TCDP \cite{tcdp}       & 80.9 \\
PointNet++ \cite{pointnet++} & 84.5 \\
PointCNN \cite{PointCNN}  & 85.1 \\
A-CNN \cite{A-CNN}      & 85.4 \\
PointWeb \cite{pointweb}  & 85.9 \\
\hline
Point2Node & {\bfseries 86.3}\\
\hline
\end{tabular}
\caption{Results on ScanNet dataset.}
\label{scannet}
\end{table}

\subsection{ScanNet Semantic Voxel Labeling}

The ScanNet \cite{scannet} dataset contains over 1,500 scanned and reconstructed indoor scenes, which consist of 21 semantic categories. The dataset provides a 1,201/312 scene split for training and testing. We followed the same data pre-processing strategies as with S3DIS, and reported the per-voxel accuracy (OA) as evaluation metrics. Table \ref{scannet} shows the comparisons between our Point2Node and other competitive methods. Our method achieves the state-of-the-art performances, due to its more effective feature learning.

\begin{table}[htbp]
\centering
\resizebox{.95\columnwidth}{!}
{
\begin{tabular}{l|c c c}
\hline
Methods    & input & \#points & OA \\
\hline
PointNet \cite{pointnet}   & $xyz$ & $1k$ & 89.2 \\
SCN \cite{scnet}       & $xyz$ & $1k$ & 90.0 \\
PointNet++ \cite{pointnet++} & $xyz$ & $1k$ & 90.7 \\
KCNet \cite{KCNet}     & $xyz$ & $1k$ & 91.0 \\
PointCNN \cite{PointCNN}  & $xyz$ & $1k$ & 92.2 \\
DGCNN \cite{dgcnn}      & $xyz$ & $1k$ & 92.2 \\
PCNN  \cite{pcnn}     & $xyz$ & $1k$ & 92.3 \\
Point2Sequence \cite{Point2Sequence} & $xyz$ & $1k$ & 92.6 \\
A-CNN \cite{A-CNN} & $xyz$ & $1k$ & 92.6 \\
Point2Node & $xyz$ & $1k$ & {\bfseries 93.0} \\
SO-Net \cite{sonet}     & $xyz$ & $2k$ & 90.9 \\
\hline
PointNet++ \cite{pointnet++} & $xyz,normal$ & $5k$ & 91.9 \\
PointWeb \cite{pointweb}  & $xyz,normal$ & $1k$ & 92.3 \\
PointConv \cite{pointconv}  & $xyz,normal$ & $1k$ & 92.5 \\
SpiderCNN \cite{SpiderCNN} & $xyz,normal$ & $5k$ & 92.4 \\
SO-Net \cite{sonet}     & $xyz,normal$ & $5k$ & 93.4 \\
\hline
\end{tabular}
}
\caption{Shape classification results on ModelNet40 dataset.}
\label{modelnet}
\end{table}

\subsection{ModelNet40 Shape Classification}
The ModelNet40 \cite{modelnet} dataset is created using 12,311 3D mesh models from 40 man-made object categories, with an official 9,843/2,468 split for training and testing. Following PointCNN \cite{PointCNN}, we uniformly sampled 1,024 points from each mesh without normal information to train our model. For classification, we removed the decoder in the Point2Node model and reported the overall accuracy (OA) as the evaluation metric.

To make a fair comparison with previous approaches that may take different information as input, we list detailed experiment setting and results in Table \ref{modelnet}. Point2Node achieves the highest accuracy among the $1k$-$xyz$-input methods. Furthermore, our $1k$-$xyz$-input model outperforms most additional-input methods, even if performing slightly worse than the best additional-input method SO-Net \cite{sonet}. Experimental results also demonstrate the effectiveness of our model in point cloud understanding.

\section{Conclusions}
We present a novel framework, \emph{Point2Node}, for correlation learning among 3D points. 
Our Point2Node dynamically enhances the nodes representation capability by exploiting correlation of different levels and adaptively aggregates more discriminate features to better understand point cloud.  
Ablation studies show that our performance gain comes from our new correlation learning modules. Results from a series of experiments demonstrate the effectiveness and generalization ability of our method.

\emph{Limitations.}
The correlation learning in Point2Node is based on refining the features (currently PointCNN~\cite{PointCNN} is used) extracted on nodes. 
While it can be considered as a general feature enhancement tool, the final performance is also limited by the capability of the pre-determined features. 
An ideal strategy would be to integrate the feature learning and correlation learning into a joint pipeline, so that the inter-channel and inter-node correlations could impact the feature extraction.
We will explore the design of such an end-to-end network in the future.

\section{Acknowledgments}
This work was supported by the National Natural Science Foundation of China (Grants No. 61771413, and U1605254).

\bibliographystyle{aaai}
\bibliography{5021-References.bib}

\section{Appendix}
\label{Appendix}

\subsection{A. Relationship to standard Non-local operations}
\label{Appendix A}
Standard Non-Local operations \cite{non-local} are widely employed in recent literature to capture long-range dependencies in 2D images and videos. To capture correlation among pixels in videos, these operations map input features with 3D convolution, and construct adjacency matrix with reshape operation and matrix multiplication. 

In our papers, to capture the correlation among nodes, we generalize standard Non-Local operations to the graph domain, where the adjacency matrix is more naturally applied for weighting edges. We not only employ Non-local operations on non-local nodes to capture long-range dependencies, but also extend them to neighboring nodes to encode local contextual information. 

Moreover, our computation is less expensive. 
In \cite{non-local}, the dimension for video data fed to the network is $T \times H \times W \times C$, where $T$ is the time of videos, $H$ is the height of videos, $W$ is the width of videos, and $C$ is the channel of videos.
The dimension of our node tensor is $N\times C$, where $N$ is the number of nodes and $C$ is the channel of nodes.
Because the 3D convolutions are replaced with 1D convolutions, and the frequent ``reshape'' operations are no needed, our computation is much more efficient.

\subsection{B. Proof of Permutation Equivariance of Dynamic Nodes Correlation}
\label{Appendix B}

Following \cite{pat}, we prove the permutation equivariance of our proposed modules, and provide the proof process as follows.

\textbf{Lemma 1.} (Permutation matrix and permutation function). \emph{$\forall A \in R^{N \times N}$, $\forall$ permutation matrix $P$ of size $N$, $\exists p:\{1,2,...,N\}\to \{1,2,...,N\}$ is a permutation function:}
\begin{equation}
    A_{ij}={(P \cdot A)}_{p(i)j}={(A \cdot P^T)}_{ip(j)}={(P \cdot A \cdot P^T)}_{p(i)p(j)}
    \label{eq17}
\end{equation}
\textbf{Lemma 2.} \emph{Given $A \in R^{N \times N}$, $\forall$ permutation matrix $P$ of size $N$,}
\begin{equation}
    softmax(P \cdot A \cdot P^T)=P \cdot softmax(A)\cdot P^T
    \label{eq18}
\end{equation}
\noindent \emph{Proof.} \\${softmax(A)}_{ij}=\frac{e^{A_{ij}}}{\sum_{n=1}^{N} e^{A_{ij}}}$.
\emph{Consider a permutation function $p$ for $P$, using Eq. \ref{eq17}, we get:}
\begin{align*}
    {(P \cdot softmax(A)\cdot P^T)}_{p(i)p(j)}&={softmax(A)}_{ij}\\
    &=\frac{e^{A_{ij}}}{\sum_{n=1}^{N} e^{A_{ij}}}\\
    &=\frac{e^{{(PAP^T)}_{p(i)p(j)}}}{\sum _{n=1}^{N}e^{{(PAP^T)}_{p(i)p(j)}}}\\
    &={softmax(P \cdot A \cdot P^T)}_{p(i)p(j)}
\end{align*}
\noindent \emph{which implies,}
\begin{align*}
    P \cdot softmax(A)\cdot P^T=softmax(P \cdot A \cdot P^T)
\end{align*}

\noindent \emph{similarly,}
\begin{equation}
     P \cdot softmax(A)=softmax(P \cdot A)
     \label{eq19}
\end{equation}
\begin{equation}
    softmax(A)\cdot P^T=softmax(A \cdot P^T)
    \label{eq20}
\end{equation}

\noindent \textbf{Proposition 1.} \emph{Self Correlation (SC) operation is permutation-equivariant, i.e., given input $X \in R^{N \times C}$, $\forall$ permutation matrix $P$ of size $N$,}
\begin{align*}
    SC(P\cdot X)=P\cdot SC(X)
\end{align*}
\noindent \emph{Proof.}
\begin{align*}
    SC(X)=X+\alpha (softmax(MLP(X))\odot X)
\end{align*}
\emph{where MLP is a node-wise function that shares weights among all the nodes,}
\begin{align*}
   MLP(P\cdot X)=P\cdot MLP(X)
\end{align*}
\emph{using Eq.  \ref{eq19}, we get:}
\begin{align*}
\begin{split}
    SC(P\cdot X)&=P \cdot X+\alpha (softmax(MLP(P \cdot X))\odot \\
    &(P\cdot X))\\
    &=P \cdot X+\alpha (softmax(P \cdot MLP(X))\odot \\
    &(P\cdot X))\\
    &=P \cdot X+\alpha (P \cdot softmax(MLP(X))\odot \\
    &(P\cdot X))\\
    &=P \cdot X+P \cdot \alpha (softmax(MLP(X))\odot X)\\
    &=P\cdot(X+\alpha (softmax(MLP(X))\odot X))\\
    &=P\cdot SC(X)
\end{split}
\end{align*}
\noindent \textbf{Proposition 2.} \emph{Local Correlation (LC) operation is permutation-equivariant, i.e., given input $X \in R^{N \times C}$, $\forall$ permutation matrix $P$ of size $N$,}
\begin{align*}
    LC(P\cdot X)=P\cdot LC(X)
\end{align*}
\noindent \emph{Proof.}
\begin{align*}
\begin{split}
    LC(X)&=LNMP(softmax(\theta ^{(l)}(DKNN(X))\cdot \\
    &(\varphi ^{(l)}(DKNN(X)))^T)\cdot DKNN(X))
\end{split}
\end{align*}
\emph{where Dilated K Nearest Neighbors (DKNN) operation is a node-wise operation,}
\begin{align*}
   DKNN(P\cdot X)=P\cdot DKNN(X)
\end{align*}
\emph{ $\theta ^{(l)}$ and $\varphi ^{(l)}$ are both node-wise functions that share weights among all the nodes,}
\begin{align*}
   \theta ^{(l)}(P\cdot X)&=P\cdot \theta ^{(l)}(X)\\
   \varphi ^{(l)}(P\cdot X)&=P\cdot \varphi ^{(l)}(X)
\end{align*}
\emph{Local Nodes Max-Pooling (LNMP) operation is invariant to local nodes' order, and permutation-equivariant to input order,}
\begin{align*}
   LNMP(P\cdot X)=P\cdot LNMP(X)
\end{align*}
\emph{using Eq.  \ref{eq18}, we get:}
\begin{align*}
\begin{split}
    LC(P\cdot X)&=LNMP(softmax(\theta ^{(l)}(DKNN(P\cdot X))\cdot \\
    &(\varphi ^{(l)}(DKNN(P\cdot X)))^T)\cdot DKNN(P\cdot X))\\
    &=LNMP(softmax(\theta ^{(l)}(P\cdot DKNN(X))\cdot \\
    &(\varphi ^{(l)}(P\cdot DKNN(X)))^T)\cdot P\cdot DKNN(X))\\
    &=LNMP(softmax(P\cdot \theta ^{(l)}(DKNN(X))\cdot \\
    &(\varphi ^{(l)}(DKNN(X)))^T \cdot P^T)\cdot P\cdot DKNN(X))\\
    &=LNMP(P\cdot softmax(\theta ^{(l)}(DKNN(X))\cdot \\
    &(\varphi ^{(l)}(DKNN(X)))^T)\cdot P^T \cdot P\cdot DKNN(X))\\
    &=LNMP(P\cdot softmax(\theta ^{(l)}(DKNN(X))\cdot \\
    &(\varphi ^{(l)}(DKNN(X)))^T)\cdot DKNN(X))\\
    &=P\cdot LNMP(softmax(\theta ^{(l)}(DKNN(X))\cdot \\
    &(\varphi ^{(l)}(DKNN(X)))^T)\cdot DKNN(X))\\
    &=P\cdot LC(X)
\end{split}
\end{align*}
\noindent \textbf{Proposition 3.} \emph{Non-local Correlation (NLC) operation is permutation-equivariant, i.e., given input $X \in R^{N \times C}$, $\forall$ permutation matrix $P$ of size $N$,}
\begin{align*}
    NLC(P\cdot X)=P\cdot NLC(X)
\end{align*}
\noindent \emph{Proof.}
\begin{align*}
    NLC(X)=softmax(\theta ^{(g)}(X)\cdot (\varphi ^{(g)}(X))^T)\cdot X
\end{align*}
\emph{where $\theta ^{(g)}$ and $\varphi ^{(g)}$ are both node-wise functions that share weights among all the nodes,}
\begin{align*}
   \theta ^{(g)}(P\cdot X)&=P\cdot \theta ^{(g)}(X)\\
   \varphi ^{(g)}(P\cdot X)&=P\cdot \varphi ^{(g)}(X)
\end{align*}
\emph{using Eq.  \ref{eq18}, we get:}
\begin{align*}
\begin{split}
        NLC(P\cdot X)&=softmax(\theta ^{(g)}(P\cdot X)\cdot (\varphi ^{(g)}(P\cdot X))^T)\cdot \\
        &(P\cdot X)\\
        &=softmax(P\cdot \theta ^{(g)}(X)\cdot (\varphi ^{(g)}(X))^T\cdot P^T)\cdot \\
        &(P\cdot X)\\
        &=P\cdot softmax( \theta ^{(g)}(X)\cdot (\varphi ^{(g)}(X))^T)\cdot P^T \cdot \\
        &(P\cdot X)\\
        &=P\cdot softmax( \theta ^{(g)}(X)\cdot (\varphi ^{(g)}(X))^T)\cdot X\\
        &=P \cdot NLC(X)
\end{split}
\end{align*}
\subsection{C. Proof of Permutation Equivariance of Adaptive Feature Aggregation}
\label{Appendix C}
\noindent \textbf{Proposition 1.} \emph{Adaptive Feature Aggregation (AFA) operation is permutation-equivariant, i.e., given input $X \in R^{N \times C}$ and $Y \in R^{N \times C}$, $\forall$ permutation matrix $P$ of size $N$,}
\begin{align*}
    AFA(P\cdot X,P\cdot Y)=P\cdot AFA(X,Y)
\end{align*}
\noindent \emph{Proof.}
\begin{align*}
\begin{split}
    AFA(X,Y)&={Gate(GNP(X,Y))}_X \odot X\\
    &+{Gate(GNP(X,Y))}_Y \odot Y
\end{split}
\end{align*}
\emph{where Global Node Pooling (GNP) operation is invariant to input order,}
\begin{align*}
    GNP(P\cdot X,P\cdot Y)=GNP(X,Y)
\end{align*}
\emph{Gate operation is shared among all the points,}
\begin{align*}
    {Gate(A)}_{P\cdot X}={Gate(A)}_X
\end{align*}
\emph{in this way,}
\begin{align*}
\begin{split}
    &AFA(P\cdot X,P\cdot Y)\\
    &={Gate(GNP(P\cdot X,P\cdot Y))}_{P\cdot X} \odot (P\cdot X)\\
    &+{Gate(GNP(P\cdot X,P\cdot Y))}_{P\cdot Y} \odot (P\cdot Y)\\
    &={Gate(GNP(X,Y))}_{P\cdot X} \odot (P\cdot X)\\
    &+{Gate(GNP(X,Y))}_{P\cdot Y} \odot (P\cdot Y)\\
    &={Gate(GNP(X,Y))}_{X} \odot (P\cdot X)\\
    &+{Gate(GNP(X,Y))}_{Y} \odot (P\cdot Y)\\
    &=P \cdot ({Gate(GNP(X,Y))}_X \odot X\\
    &+{Gate(GNP(X,Y))}_Y \odot Y)\\
    &=P \cdot AFA(X,Y)
\end{split}
\end{align*}

\end{document}